\def\year{2017}\relax
\documentclass[letterpaper]{article}
\usepackage[noblocks]{authblk}
\usepackage{aaai17}
\usepackage{times}
\usepackage{helvet}
\usepackage{courier}
\usepackage{url}
\usepackage{graphicx}
\usepackage[T1]{fontenc}

\begin{document}
\title{Laying Down the Yellow Brick Road: Development of a Wizard-of-Oz Interface for Collecting Human-Robot Dialogue}
\author[1]{Claire Bonial}
\author[1]{Matthew Marge}
\author[2]{Ron Artstein}
\author[1]{Ashley Foots}
\author[3]{Felix Gervits}
\author[1]{Cory J. Hayes}
\author[1]{Cassidy Henry}
\author[1]{\authorcr Susan G. Hill}
\author[2]{Anton Leuski}
\author[1]{Stephanie M. Lukin}
\author[1]{Pooja Moolchandani}
\author[1]{ Kimberly A. Pollard}
\author[2]{ David Traum}
\author[1]{Clare R. Voss} 

\affil[1]{U.S. Army Research Laboratory, Adelphi, MD 20783}
\affil[2]{USC Institute for Creative Technologies, Playa Vista, CA 90094}
\affil[3]{Tufts University, Medford, MA 02155}
\affil[ ]{\makebox[0pt]{\ttfamily claire.n.bonial.civ@mail.mil}%
  \makebox[0pt]{\raisebox{2in}[0pt][0pt]{Published in the 2017 AAAI
      Fall Symposium on Natural Communication for Human-Robot Collaboration}}%
  \makebox[0pt]{\raisebox{1.8in}[0pt][0pt]{https://www.aaai.org/ocs/index.php/FSS/FSS17}}}
\maketitle
\begin{abstract}
We describe the adaptation and refinement of a graphical user interface 
designed to facilitate a Wizard-of-Oz (WoZ) approach to collecting human-robot 
dialogue data. The data collected will be used to develop a
dialogue system for robot navigation.
Building on an interface previously 
used in the development of dialogue systems for virtual agents and
video playback,
we add templates with open parameters which allow the wizard to
quickly produce a wide variety of utterances.
Our research demonstrates that this 
approach to data collection is viable as an intermediate step in  
developing a dialogue system for physical robots in remote locations from their users~-- a domain in which 
the human and robot need to regularly verify and update a shared 
understanding of the physical environment.  
We show that our WoZ interface and the fixed set of utterances 
and templates therein 
provide for a natural pace of dialogue with
good coverage of the navigation domain.
\end{abstract}

\section{Introduction}

For robots to become effective teammates with humans at collaborative tasks 
such as search-and-rescue operations and reconnaissance, 
they must be able to communicate effectively with humans 
in dynamic environments.  Ideally, these robot collaborators could 
engage in two-way spoken dialogue, which 
is both natural for humans and
efficient for exchanging information 
about tasking, goals, situational awareness, and status updates.

To develop a robot's dialogue capabilities, we need data on how people
might talk to that robot.
We collect these data through
the Wizard-of-Oz (WoZ) methodology \cite{Dahlback-wizard}, used in
virtual human dialogue systems 
to refine and evaluate the domain and provide training data for
automated natural language understanding \cite{traum_dealing_2005,devault2014simsensei}.
A critical research 
question is whether or not this virtual-agent approach can be extended to and effective 
for the physically grounded, situated language needed for 
communication with a robot in a collaborative task, especially if the human and robot
are not co-present.
Thus far, our results support the viability 
of
this approach
for our scenario.  In this paper, we describe 
our progress, with particular attention paid to the adaptation and refinement of the 
wizard user interface for the research phase of collecting dialogues for training data.  

The importance of the interface is twofold.  
First, it
eases the physical and cognitive overhead of the
wizard (compared to manual typing).
Second, decisions on the communications built into 
the interface represent a critical research step:
mapping the unconstrained language of a na\"ive participant 
into a set of communication intents that can be understood and 
acted upon by a dialogue system.
The mappings, as collected in experiments,
then serve to train an automated dialogue system.
The interface limits communications to a fixed set, 
yet it must provide adequate 
coverage for communicating about the tasks, environment, and 
overcoming miscommunication.  
Our results show that: (i)~the interface 
facilitates a faster pace of communication that 
approaches more natural dialogue exchanges and (ii)~the set 
of utterances and templates it contains
provides good coverage 
of the domain.

\section{Related Work}

The WoZ method involves one or more
human ``wizards'' standing in as AI modules, performing functions
that will eventually be performed by a final automated system.
This behind-the-scenes human activity is 
unknown to the research participants (so long as the WoZ 
illusion is successful).
The WoZ methodology is useful due
to its low development cost when
technology to support the desired
functionality does not yet exist. 
In the human-robot interaction domain, natural
language interpretation is one of the most common
use cases for WoZ~\cite{riek2012wizard}.  
It has traditionally
been used as a surrogate for automatic speech
recognition~\cite{zollo1999study,skantze2003exploring},
and to train
a dialogue 
system by progressively adding automation
over several development stages~\cite{passonneau2011embedded}.

Our setup relies on two wizards, one
for dialogue management and one for
robot navigation, 
as a result of prior pilot trials 
when a single wizard struggled to perform both functions~\cite{cassidy2015turn}. 
This setup is
similar to the SimSensei
project~\cite{devault2014simsensei}, where during the data collection phase, two 
wizards stood in for what would ultimately
be separate software components 
(i.e., verbal and non-verbal
behaviors of the virtual agent). 
Green et al.~\shortcite{green2004applying} 
investigated the use of multiple wizards
for dialogue processing and navigation capabilities
for a robot in a home touring scenario. This work
found the multi-wizard approach to be valid in situations
where the robot and human were co-present. 
We expand on this method by
addressing remote (not co-present) human-robot communication.

Other research focused on developing an adaptable WoZ interface like our interface includes 
the SUEDE tool \cite{klemmer2000suede} and DOMER interface \cite{villano2011domer}, 
both of which use a 
similar development strategy to ours, with 
 iterative expansion and refinement of the responses to be
 included in the interface.

\section{Background \& Approach}

The long-term vision of our work is to provide more 
natural ways for humans to interact
with robots in shared tasks.
The WoZ methodology facilitates a data-driven understanding of how 
people talk to robots in our collaborative domain.  Similar to 
\citeauthor{devault2014simsensei}\ \shortcite{devault2014simsensei}, we use the WoZ methodology only in the 
early stages of a multi-stage development process 
in order to refine and evaluate the domain and provide training data for
automated dialogue system components.  In all stages of this process, 
participants 
speak freely,
even as increasing levels of automation are introduced in each subsequent
stage or ``experiment,'' using
data from previous experiments.

\begin{figure}[t!]
\centering
\includegraphics[width=3.2in]{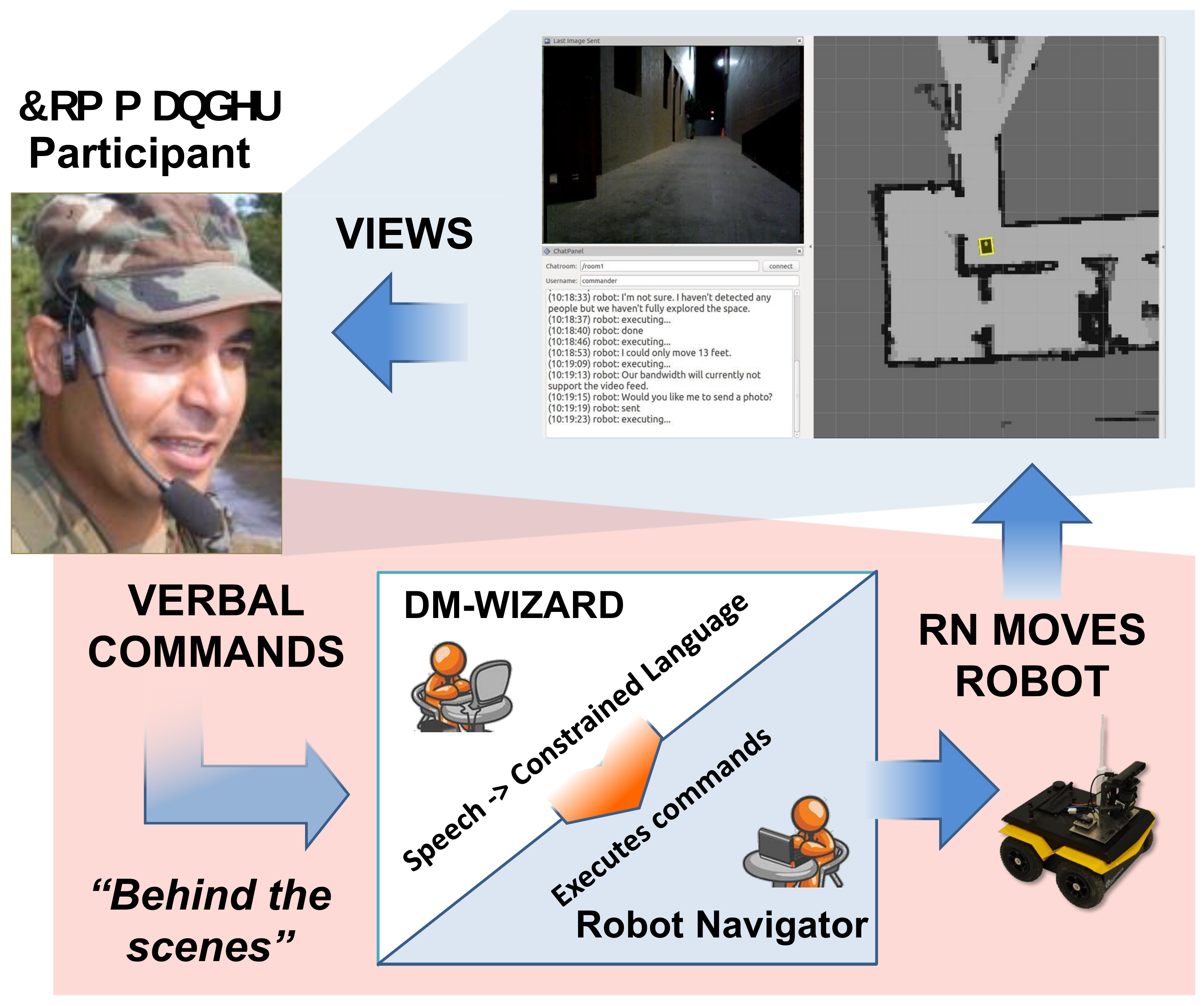}
\caption{WoZ setup~\cite{roman2016-wizards}.} 
\label{expt-layout}
\end{figure}

The first two experiments on the path to increased automation use two wizards: a Dialogue Manager 
Wizard (DM-Wizard) who sends text messages and a Robot Navigator Wizard  (RN-Wizard) who teleoperates the robot. 
A na\"ive participant is 
tasked with instructing a robot to navigate through a remote, unfamiliar house-like 
environment.  The participant is seated at a workstation equipped with a 
microphone and a desktop computer
displaying
information collected by the robot: a map 
of the robot's position and its heading in the form of a 2D 
occupancy grid, the last still-image captured by the 
robot's front-facing camera, and a chat window showing text 
responses from ``the robot.'' This
layout is shown in the 
top, right-hand corner of  Figure~\ref{expt-layout}, which represents 
an overview of our WoZ setup.

At the beginning of the study, the participant receives a list
of robot capabilities: the robot understands basic object properties
(e.g., most object labels, color, size), relative proximity, some spatial terms, and
location history.  
Experimenters do not give example
instructions, but rather tell 
the participant that s/he can communicate in spoken 
language
using natural expressions to complete tasks. 
In reality, 
the participant is speaking not to a robot, 
but 
to an unseen DM-Wizard  
who listens to the participant's spoken 
instructions and responds with text messages in a chat window.  
There are two high-level response options: 

1) If the 
instructions are clear and executable in the current physical 
environment, then the DM-Wizard passes a simplified text version 
of the instructions to the RN-Wizard, who 
then joysticks the robot to complete the instructions. 

2) If the instructions are problematic in some way, due to 
ambiguity or impossibility given either the current physical context
or the robot's capabilities, then 
the DM-Wizard 
responds directly to the participant in text via a chat window,
in order to clarify the instructions and/or 
correct the participant's understanding of the robot's capabilities.

We engage each participant in three sessions: a 
training task and two main tasks.  The two 
main tasks, lasting 20 minutes each, focus on 
slightly different search and analysis subtasks and start in distinct locations 
within a house-like environment.  
The subtasks were developed to encourage the
participant to treat the robot as a teammate who helps search for certain objects, 
but also to encourage the participant to tap into their own
real-world knowledge to analyze the environment. 
An example search task was to locate
\textit{shoes}, relying on robot-provided
images. An example analysis task was to consider whether the 
explored space was suitable as a headquarters-like environment.

The human-robot communication data collected in Experiments (Exp) 1 and 2 
allow us to incrementally build  
a fully automated dialogue system on-board a physical robot. 
In Exp 1, our goal was to elicit the full range 
of communications that may arise in our domain.  To allow for this, 
the DM-Wizard simply typed
free 
responses to the participant following guidelines
 established during piloting that governed the DM-Wizard's 
real-time decision-making \cite{roman2016-wizards}.
The guidelines identified the minimal requirements
for an executable instruction: each must contain both a clear
action and respective endpoint. 
The guidelines also provided response categories and templates,
giving the DM-Wizard easily-remembered templates for elements of each response, 
but also flexibility in exact word choice.
Data from ten participants was collected in Exp 1.

The Exp 1 data was then analyzed to develop a set of communications to design into an interface for Exp 2, where messages sent via that interface would strike a balance 
between tractability for an automatic system and full coverage of the 
domain, including recovering from problematic instructions.  
With the interface, 
instead of typing free responses, the DM-Wizard constructs a response 
by selecting buttons on a graphical user interface (GUI), where each button press 
sends a text message either to the participant or the RN-Wizard. 
Aside from the DM-Wizard communicating via the interface,
Exp 2 was conducted just like Exp 1.
Ten new participants took part in Exp 2.

Before delving into the development of this interface that preceded running Exp 2, 
we should note that we have already completed Exp 2 and used the collected data to train a classifier 
that automatically generates responses to particular participant instructions. 
As a result, it is quite clear that additional training data is needed for a robust classifier
and so in the planned Exp 3, we are 
simulating both the physical environment and the robot \cite{cassidy2017towards}.  The simulation
 will allow us to reduce the time and space overhead otherwise needed for 
performing this experiment using a physical robot in a real environment; 
therefore speeding up the collection of training data.
As we collect adequate levels of training data, in future experiments we 
will begin to automate away individual components, currently using wizard stand-ins. 
For further discussion of this development process, see Ongoing \& Future Work.

\section{Interface Development}

The DM-Wizard interface, which is the key change in the 
 setup between Exps 1 and 2, 
sends a text response to either the participant or the RN-Wizard.  
The critical research challenge in developing this interface has been to capture
 the sum total of possible  
responses ``the robot'' can give to the RN-Wizard and participant, whose language 
is totally unconstrained and can vary widely.  
Thus, the quality of the trained model is contingent upon the GUI design decisions. 
 However, the goal of domain coverage had to be balanced with the need to 
create an interface that was organized such that the wizard could easily 
and quickly find the appropriate button -- presenting another development challenge. 
 
\subsection{Software Overview}
The interface software
was adapted from a design used for WoZ prototyping of a dialogue system in which 
humans can engage in \textit{time-offset interaction} with a WWII Holocaust 
survivor \cite{artstein2015many}. In that application, people could ask 
a question of the system, and a pre-recorded video of the Holocaust 
survivor would be presented, answering the question.  The 
interface is implemented as a web application that presents a collection 
of clickable buttons in a web browser window.  In our study, the DM-Wizard uses the 
interface buttons to trigger a text response to be sent to the appropriate 
chat window in either the participant's screen or the RN-Wizard's 
screen.  The system uses the VHMsg messaging protocol \cite{hartholt2013all}
 built on top of the ActiveMQ message broker.\footnote{\url{http://activemq.apache.org}}
 The rest of the system components, including tools for logging, data visualization, and robot operation, interact via ROS (Robot Operating 
 System).\footnote{\url{http://www.ros.org}} We also
 implemented a software bridge that connects to VHMsg and the ROS message server and maps VHMsg messages onto ROS messages and vice versa automatically.

\begin{figure}[t!]
\centering
\includegraphics[width=3in]{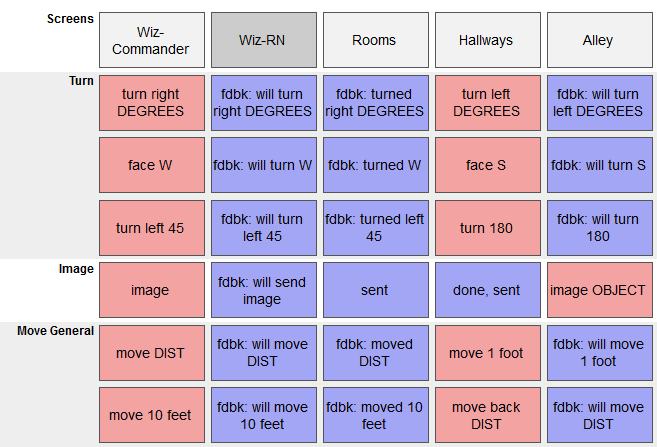}
\caption{DM-Wizard interface with Wiz-RN tab displayed: red buttons pass messages to RN-Wizard, blue to participant; all caps words 
indicate text-input slots labeled by input type.} 
\label{dm-gui}
\end{figure}

There are a large number of responses, and therefore buttons, needed 
in the interface to provide coverage for all of the participant
instructions that must be passed to the RN-Wizard, and for all of the possible 
responses needed to clarify or acknowledge
instructions.
To
organize the large number of buttons (that do not fit in a single screen), 
there are tabs to switch between five screens, which present thematically related 
buttons.  Within each screen, there are labeled rows of buttons, which represent 
subthemes.  As needed, a very frequently used button (e.g., ``done'') may appear in more than one location to
speed up the DM-Wizard's ability to find it.  Color-coding and a short 
label on the button aid in quick identification, and the full content of the message 
associated with a button can be viewed by hovering over it. A snippet of the interface is shown 
in Figure~\ref{dm-gui}.

The initial WoZ design assumed that the message templates are static and can be fully configured before running the system. In our preliminary analysis, we observed that a significant proportion of DM-Wizard messages fell into well-defined patterns or templates, e.g., ``I see a door on the left,'' 
``I see a door on the right,'' ``I see a wall,'' etc.
In such cases, we observed that it would be difficult to enumerate all but the most frequently occurring objects in the scenario. 
Thus, we extended the WoZ interface to allow the 
DM-Wizard to modify the message content on the fly. 
Specifically, a button message text may now 
include a text-input slot. When the 
DM-Wizard clicks on such a button,
a pop-up window appears with a text-input field, e.g., ``I see \_\_\_.'' The DM-Wizard types the object description and
sends the newly completed
message to its recipient. Note that there 
is no entirely open response button, all buttons reflect, at a minimum, an observed template of responses 
like ``I see \_\_\_''.  Other text-input slots are shown in all caps in Figure~\ref{dm-gui}.

We used an iterative approach to develop  
the final interface. 
This was carried out by testing 
it in practice runs in which we replayed 
audio from Exp 1, followed by a round of
real-time Exp 2 pre-piloting.
The refinements can be broadly categorized 
 into
changes that affected the 
content of the interface (i.e. the messages associated with
 buttons) and changes in the layout of the buttons.

\subsection{Content}
The finite set of options included in the DM-Wizard interface 
were carefully selected to strike a balance between 
providing enough expressive options for the DM-Wizard to pass 
along executable instructions and overcome 
communication problems, and limiting the options to a set 
that is tractable for automation.  Although the WoZ methodology 
using a fixed set of messages has been 
shown to be effective for communication with virtual agents, 
the adaptation of this approach to human-robot dialogue explored 
a new research question: whether this approach is viable for 
situated language, where a shared understanding of the physical 
environment, including objects, locations, paths and orientations, is needed, particularly
when the human and robot are not co-present.  

\textbf{Initial Content} -- To gain a preliminary understanding of the messages 
needed in the interface, we undertook thorough analysis and annotation 
of the Exp 1 data, in which the DM-Wizard had typed relatively free responses 
to the participant.  
The DM-Wizard responses were categorized first according to their 
recipient (either the participant or the RN-Wizard), and then by their 
function (e.g., \textit{feedback}, \textit{clarify-distance}, 
\textit{describe-capability}) \cite{marge2016assessing}.  These annotations 
were used to create the first version of the interface, in which an effort was made to ensure 
that each DM-Wizard message 
 occurring 2 or more 
times in Exp 1 was captured in the interface.  For communications 
that occurred very frequently, alternatives were added with slightly 
different lexical choices.  These 
can be alternated as needed to reduce the potentially unnatural, repetitive 
feel of the robot's communications.  

\textbf{Metric References} -- While adding the appropriate button was straightforward for many of 
the common messages sent to the participant (e.g., ``I'm not sure which 
object you are referring to,'' ``Executing,'' ``Done''), this process 
was not as straightforward for instruction messages sent to the RN-Wizard, which 
regularly drew upon the physically grounded, situated nature of the 
language in this domain.
 Common instructions include asking the robot to move forward a particular number of 
feet, rotate left or right a particular number of degrees, and moving to an 
object or landmark in the physical surroundings \cite{marge2017exploring}. 
Despite these commonalities, providing coverage for these instructions 
in the interface was challenging given that the specific number of feet 
(or another unit of measurement) and degrees would vary.  In lieu of 
including a button for every possible unit of movement, we took advantage of 
the interface capability for buttons with text-input fields, which allow the DM-Wizard 
to type in the specific number of feet or degrees requested.  Because 
use of this type of button is more time-consuming, very common measurements 
found in the instructions were given their own, dedicated button (e.g., 
``turn left 90 degrees,'' ``move forward two feet''). Nonetheless, having 
some flexibility in certain common instructions allows for the generalization needed 
to avoid an over-proliferation of buttons that would be unmanageable 
for the DM-Wizard.   

\textbf{Location \& Object References} -- Although it was also 
possible to use buttons with text-input fields to cite specific objects and landmarks 
used in instructions, we decided to create a map of our physical environment 
and assign unique identifiers to all spaces (i.e. rooms, hallways), objects, 
and doorways.  We then added buttons accommodating 
instructions to move to/through/parallel to each doorway in the environment, 
move to each object in the environment, and move into 
each specific room and hallway space in the environment.  Thus, for example, 
if a participant instructed the robot to ``go ahead to the office/black/padded chair ahead,'' 
this would be translated to an RN-Wizard instruction using a specific identifier,  
e.g., ``move to Conference Room Right Chair 1.'' Although this more 
than doubled the number of buttons in the interface, our hope was that this would make 
our training data more tractable for a preliminary system in which the environment 
is known, and particular locations and objects are assigned coordinates 
in that environment.  To make the larger number of buttons more manageable, 
instructions using unique identifiers associated with different spaces in the environment 
were separated out into different tabs/screens (see top row of tabs in Figure~\ref{dm-gui}).

\textbf{Feedback Messages} -- In Exp 1
participants remarked on the slow pace of the experiment.  
Even in cases of entirely successful communications 
and executable instructions, the relatively slow pacing of the dialogue 
could lead the participant to believe that something 
had gone wrong.  The introduction of an interface in Exp 2 partially addresses this problem 
simply because it allows the DM-Wizard to pass messages more quickly without 
typing.  We also wanted to provide more transparency to participants 
as to what the 
robot had understood and what it was doing, ensuring they could be confident 
in the success of their communications.  
Thus, a major area of content additions to the interface 
were different types of feedback and back-channeling to improve the 
transparency of what the robot was doing \cite{Allwood92} and help to establish common
ground~\cite{Clark89}.  

In Exp 1, the robot 
would often only respond with ``Executing'' when a set of instructions was underway 
 and ``Done'' 
when instructions were completed.  In Exp 2, the GUI allowed us to make this 
feedback much more nuanced.  For clear cases of executable instructions, 
the DM-Wizard would immediately acknowledge that the instructions were heard and 
understood by responding 
with ``ok'' followed by a demonstration of that understanding in a repetition or 
paraphrasing of the instructions given.  This repetition was either in the form of a description 
of what the robot was about to do, or a description of what the robot just did: ``I'll move forward three feet,'' 
or ``I moved forward three feet.''
On the other extreme, in cases where 
the executability of the instructions was very uncertain and the DM-Wizard needed time 
to consider the surrounding physical environment, the immediate response was ``Hmmm...,'' 
which could be followed either by a confirmation of the intention to complete 
the instructions or a description of the problem and a suggestion for an alternate, 
potentially helpful action (e.g., ``There's an obstruction preventing me from doing that.  Would you
like me to send a picture?''). We can anecdotally report that these feedback additions seemed to help the robot 
to hold the floor when instructions were being completed -- participants seemed less likely 
to assume something had gone wrong and abandon one strategy of instructions in favor 
of a new strategy 
(quantitative verification of this observation is underway).

\textbf{Decomposing Complex Instructions} -- Another area of major content changes was the decomposition 
of complex instructions to the RN-Wizard.  At this stage of our research, we have not fully fleshed out 
how the instructions passed to the RN-Wizard could be translated into policies 
executable by a robot's planning component.  Nonetheless, in the development of 
the interface, we were forced to consider what 
policies might be associated with each button that passed a particular set of 
instructions to the RN-Wizard.  For example, the interface button ``Move to Kitchen Door''  was 
originally associated with 
a policy of (i) moving to the vicinity (within one robot's length) of the referenced door 
and then (ii) orienting to face the door -- thus this was a ``complex'' instruction in the sense that 
it was associated with more than one move/turn action.  Although we had hypothesized this sequence was what 
most participants would want the robot to do given this instruction, in pre-piloting we found 
that participants often used doorways as landmarks for moving 
certain distances, but they weren't always interested in the doorway itself.  In these cases, 
they might say something like ``Move parallel to the doorway ahead on the left,'' 
an action that we couldn't execute precisely since our set of instructions always included 
a policy of both moving and orienting to face a referenced doorway.  

We discovered that the messages to the RN-Wizard could be much more flexible and composable 
if we limited the number of complex instructions (containing more than one move/turn instruction)
 included, and decomposed 
most into individual buttons/instructions associated with only one action. 
 In the case of doorways, we opted to associate 
``Move to Kitchen Door'' with moving into the vicinity (within one robot's length) of the 
referenced door only.  We then added 
buttons for each door that accommodated instructions to move parallel to each 
door and turning to face each door in the environment.  In general, 
the flexibility afforded by decomposing messages into the smallest action or intentional units 
led to similar decomposition in the DM-Wizard to participant communications.

\subsection{Layout}
During development, we changed the layout of the buttons and screens
by shifting from a thematically related layout
to one
that better accommodated which buttons were often used together in a series 
in practice.  The best example of this
allowed 
for the DM-Wizard to quickly communicate instructions to the RN-Wizard and corresponding 
feedback to the participant.  In the 
original formulation of the interface, 
one screen was dedicated for communications to the participant, and 
a second screen for communications to the RN-Wizard. 
This layout forced the DM-Wizard  
to constantly switch screens in order to first provide instructions to the RN-Wizard and then feedback to the 
participant.  In answer to this challenge, 
the screen previously dedicated to communications to the RN-Wizard was 
augmented with color-coded buttons that sent feedback to the participant.  
So, for example, next to the ``Move forward three feet''  button 
with instructions to the RN-Wizard,
we added
feedback buttons ``I will move forward three feet'' or ``I moved forward three feet'' to the participant.
Examples can be seen in Figure~\ref{dm-gui}, 
where color coding indicates the message recipient.
This 
altered  layout made it easier for the DM-Wizard to provide timely and specific feedback 
to the participant. 
In the future, we are considering the introduction of buttons that simultaneously 
send one message to the RN-Wizard, and another corresponding feedback message to the participant.

\section{Impact of Interface}

The introduction of the interface in Exp 2 impacted the dialogue data collected 
by limiting DM-Wizard utterances to the coverage available in the interface buttons, and 
by speeding up the pace of dialogue. 

\subsection{Coverage}
In development, an effort was made to ensure that the Wizard interface provided coverage for 
all DM-Wizard 
messages with two or more occurrences collected in Exp 1.
 Out of a total of 2,728 DM-Wizard messages in Exp 1, 
there were 2,075 that occurred two or more times 
with 84 unique messages covering 76\% of the total.  
This regularity is expected, and arises in part from the nature of the domain 
and task, but also from the DM-Wizard being guided by the policies and 
templates for responses given in the Exp 1 guidelines.   
Of course, the singletons that we did not accommodate with a dedicated 
response button still require some type of response, so these were handled 
through one of the following strategies: 

1) buttons with text-input fields 
accommodate less frequent measurements in common move or turn-type 
instructions; 

2) buttons with somewhat generalized vocabulary are available
for clarifications (e.g., ``the one on my left?'' provides coverage for 
a more specific response in Exp 1: ``the crate on my left?''); and 

3) buttons with very generic indications of the problematic nature of an instruction 
(e.g., ``I'm not sure 
what you are asking me to do; can you describe it another way?'') 
accommodate truly novel cases that may be off-topic, entirely outside of 
the robot's represented capabilities, or very ambiguous and/or nonsensical 
given the physical environment, (e.g., ``rotate left 200 feet'').

\noindent The final interface contained 404 buttons, organized into 5 screens, where the first two 
screens 
convey the most common recipient for buttons on that page (either the 
participant, called ``Commander,'' or the RN-Wizard), and the last three screens 
contained buttons that reference different objects/areas in the physical environment.  

After the iterations of interface refinement were completed, its coverage 
was analyzed using string matching to compare all of the DM-Wizard messages in 
Exp 1 to the messages included in the finalized interface.  We found that 88.7\% 
of the messages in Exp 1 have an exact match in the interface buttons, while 10.5\% 
have partial matches with fairly clear candidates for the best interface button, and 
0.8\% have no match and no clearly corresponding interface button.  In qualitative analysis, 
we see that the partial 
matches generally reflect cases handled with buttons with text-input fields 
or more general vocabulary (strategies (1) and (2) above), 
while the no-match cases are places where a more generic strategy of 
non-understanding would be used (strategy (3)).  See Appendix 
for examples of parallel Exp 1 and 2 DM-Wizard responses.

\subsection{Speed}
Because of the DM-Wizard's cognitive and physical overhead, the pacing of dialogue in Exp~1 
was quite slow.\footnote{Observed DM-Wizard delays
lasted well beyond 700 ms pause thresholds in spoken dialogue systems  \cite{raux2008optimizing}.} 
Although both the experimenter and ``the robot'' 
explain to the participant that ``there may be lag times,''
the slow pacing certainly affected the participant's 
perception of the robot as an interlocutor.  Schegloff (2007) notes that 
``preferred'' responses in dialogue tend to occur immediately after a single beat of 
silence that is the normal transition place from one interlocutor to another\nocite{schegloff2007sequence}. 
On the 
other hand, a common feature of ``dispreferred'' responses in dialogue is that 
they are delayed -- there is a longer period of silence prior to the response.  
Thus, human interlocutors are tuned to understand timely feedback 
as an indication that everything is going well in the interaction, while silence 
can indicate a problem.
Given these features of natural dialogue, we felt it important 
to speed up the pace of the interaction and avoid long silences where the participant 
was not receiving feedback as to what the robot was doing.  The introduction of the 
interface facilitates faster responses and, since typing isn't needed, the feedback can also be longer 
and more specific (e.g., ``Ok, I'll move forward three feet,'' as opposed to ``Executing,'' 
in Exp~1).  

Currently, we are processing the data needed to examine 
exact differences in response times between Exps 1 and 2.  However, 
we can gain a sense of the faster pace of dialogue in Exp 2 by examining 
how much the 10 participants were able to accomplish in Exp 1 
compared to Exp 2.  The time allotted (two 20-minute phases) and 
tasks in both experiments are the same; tasks 
can be broadly categorized as search and navigation tasks.  Thus, most instructions 
are either requests for the robot to move or turn (navigation), or to send 
a picture of the current environment to see if it 
includes target objects of interest (search). Anytime a move or turn action was 
successfully completed by the robot, completion feedback was given (e.g., ``done'').
Similarly, anytime a request for a picture is fulfilled, completion feedback was given in the form of ``sent.''  We can therefore report the number of times this type of completion feedback is given 
as a proxy measure for the pacing of dialogue.  
In Exp 1, across all 10 
participants, there are 829 instances of completion feedback, indicating that 829 task-oriented actions 
were completed.  In Exp 2, there are 1069 instances of completion feedback -- indicating that 
participants were able to successfully complete more subtasks in Exp 2.  Admittedly, this is 
a flawed measure, given that the participants may individually differ in how they approached the task 
and how successful they were.  Nonetheless, it does reflect a trend in Exp 2 
towards faster-pacing, which is closer to a natural human dialogue pace and 
which allows participants to complete more tasks in the allotted time.  

\section{Ongoing \& Future Work} 
\label{future}

Creating the interface was our first step in automating the robot partner
by providing a more advanced tool for the DM-Wizard's
natural language generation tasks. We plan 
to automate additional tasks, ultimately removing the wizards.

\textbf{Automating the Dialogue Manager} -- 
 We are currently
investigating use of the  
 NPCEditor
\cite{leuski2011}, which has been used
 extensively for virtual human applications (typically question-answer
 systems).  It performs natural language understanding 
by learning a mapping between input utterances and
 their associated responses. In our case, the utterance-response
 mappings will be extracted from our experimental data.

We
 expect the more structured language output from the GUI to simplify
 this process.

\textbf{Automating the Robot Navigator} --
In the current experiments, we rely on human intelligence to interpret
high-level instructions including metric information and landmarks. While
these instructions clearly specify some aspects of the endpoints, they
are often underspecified in terms of the exact path to be taken and
the final specific location and orientation. We are designing a set of policies for a 
robot planner in order to interpret
these high-level instructions and guide the robot to the appropriate
endpoints. Since the path and final positioning derived by an automated
planner might be different from that taken by a human RN-Wizard, we  may 
need additional feedback to and clarification from the
participant. We are testing variations of the observed instructions  
and paths taken in
order to provide a general action capability.

\section{Conclusions}

The progress described here demonstrates  
that WoZ approaches taken in virtual human applications 
can also be successfully adapted to this new domain 
of human-robot dialogue, 
where the interlocutors are not co-present and must rely on 
a shared understanding of the physical surroundings.  
Although we are in the early stages of our research, 
the implementation of a Wizard interface discussed here represents 
an important
milestone in establishing the feasibility of this approach.  
 Our
domain 
requires 
a great deal of flexibility in how ``the 
robot'' interacts with the physical environment and 
handles various na\"ive participants' strategies in completing 
search and navigation tasks.
We have shown that the interface provides coverage  
of the domain with a limited set of messages and templates, 
carefully crafted to generalize over distinct contexts.  
Additionally, the interface has been crucial in eliciting more natural 
dialogue since it allowed for faster-pacing of turns, to which 
humans are accustomed.
\\
\\ 

\section{Acknowledgments}
This research was sponsored by the U.S. Army Research Laboratory. 
\section{Appendix: Dialogue Samples}
The following dialogue snippets illustrate the difference between
Exp.~1 ``free'' response mode and Exp.~2 GUI-structured Response Mode.
The dialogues are between a Participant (P) and DM-Wizard (DM).
Note the increased 
consistency
of DM-Wizard responses to the open-ended
instructions ``move forward'' in Exp.~2.

\subsection{Exp.~1 Free Response Mode}

\begin{small}
\begin{description}
  \setlength{\parskip}{0pt}
  \setlength{\itemsep}{0pt}

\item[P] move forward 
\item[DM] You can tell me to move to something that
	you see. 
\item[P] go forward four feet	

\vspace{\baselineskip}

\item[P] continue to move down the hallway	 
\item[DM] How far should I continue down the hallway? 
	Would you like me to reach something that you see 
	or go a certain distance? 
\item[P] go down the end of the hallway where it appears to be a 
clock on the wall 
\end{description}
\end{small}

\subsection{Exp.~2 GUI-Structured Response Mode}

\begin{small}
\begin{description}
  \setlength{\parskip}{0pt}
  \setlength{\itemsep}{0pt}

\item[P] robot go forward	 
\item[DM] I'm not sure where or when to stop 
	moving forward. You can tell me to move a certain 
	distance or to move to an object. 
\item[P] robot move forward one feet 

\vspace{\baselineskip}

\item[P] can you move forward	
\item[DM] I'm not sure where or when to stop
	moving forward. You can tell me to move a certain 
	distance or to move to an object. 
\item[P] can you stop at the second door 
\end{description}
\end{small}

\begin{quote}
\begin{small}
\bibliographystyle{aaai}
\bibliography{aaai17-gui}
\end{small}
\end{quote}

\end{document}